%% file: main.tex
\documentclass[10pt,twocolumn,letterpaper]{article}

\usepackage[pagenumbers]{cvpr} 
\usepackage{graphicx}   
\usepackage{booktabs}   
\usepackage{multirow}   
\usepackage{listings}
\usepackage{float}  

\input{preamble}
\definecolor{cvprblue}{rgb}{0.21,0.49,0.74}
\usepackage[pagebackref,breaklinks,colorlinks,allcolors=cvprblue]{hyperref}

\def\paperID{11219} 
\def\confName{CVPR}
\def\confYear{2026}

\title{HiVid-Narrator: Hierarchical Video Narrative Generation with Scene-Primed ASR-anchored Compression}

\author{Haoxuan Li\\
Taobao \& Tmall Group of Alibaba\\
\and
Mengyan Li\\
Taobao \& Tmall Group of Alibaba\\
\and
Junjun Zheng\textsuperscript{†}\\
Taobao \& Tmall Group of Alibaba\\
}
\begin{document}
\maketitle
\input{sec/0_abstract}    
\input{sec/1_intro}
\input{sec/2_relatedwork}
\input{sec/3_method}

\input{sec/4_experiments}
\input{sec/5_conclusion}
{
    \small
    \bibliographystyle{ieeenat_fullname}
    \bibliography{main}
}

\input{sec/X_suppl}

\end{document}

%% file: sec/0_abstract.tex
\begin{abstract}
Generating structured narrations for real-world e-commerce videos requires models to perceive fine-grained visual details and organize them into coherent, high-level stories—capabilities that existing approaches struggle to unify. We introduce the \textbf{E}-commerce \textbf{H}ierarchical \textbf{V}ideo \textbf{C}aptioning (\textbf{E-HVC}) dataset with dual-granularity, temporally grounded annotations: a Temporal Chain-of-Thought that anchors event-level observations and Chapter Summary that compose them into concise, story-centric summaries. Rather than directly prompting chapters, we adopt a staged construction that first gathers reliable linguistic and visual evidence via curated ASR and frame-level descriptions, then refines coarse annotations into precise chapter boundaries and titles conditioned on the Temporal Chain-of-Thought, yielding fact-grounded, time-aligned narratives. We also observe that e-commerce videos are fast-paced and information-dense, with visual tokens dominating the input sequence. To enable efficient training while reducing input tokens, we propose the \textbf{S}cene-\textbf{P}rimed \textbf{A}SR-anchored Compressor (\textbf{SPA-Compressor}), which compresses multimodal tokens into hierarchical scene and event representations guided by ASR semantic cues. Built upon these designs, our HiVid-Narrator framework achieves superior narrative quality with fewer input tokens compared to existing methods.
\end{abstract}

%% file: sec/1_intro.tex
\section{Introduction}
E-commerce videos have transformed how consumers discover and purchase products, with platforms like Taobao and TikTok generating billions of hours of product demonstration videos annually. These videos are information-dense and fast-paced: a typical 60-second clip may showcase multiple product features, demonstrate various usage scenarios, and transition rapidly between close-ups and wide shots—all while the host provides detailed commentary. For viewers who cannot watch in real-time or prefer text summaries, automatically generating structured narratives from such videos becomes crucial for product discovery and decision-making.

However, existing video understanding approaches fall short in this domain. Traditional dense video captioning methods~\cite{krishna2017dense,zhou2018towards} generate flat, event-level descriptions without narrative coherence. Recent multimodal large language models (MLLMs)~\cite{ren2024timechat,huang2024vtimellm} have improved temporal reasoning but treat narration as a single-step generation task, lacking explicit grounding of visual details before summarization. This often leads to hallucinated product attributes or omitted key features—critical failures in e-commerce contexts where factual accuracy directly impacts purchasing decisions. Moreover, these models struggle with the computational burden of processing hundreds of densely informative frames, making them impractical for real-world deployment.

We argue that generating high-quality narrations for e-commerce videos requires a hierarchical reasoning process: models must first temporally ground fine-grained visual semantics and then synthesize them into coherent, chapter-based summaries. We formalize this as Hierarchical Video Narrative Generation aligned with our dataset design: (1) Temporal Chain-of-Thought, a time-stamped, segment-by-segment description of on-screen content, actions, and attributes; and (2) Chapter Summary, contiguous chapters each with a title and concise summary. This mirrors human annotation: careful temporal inspection followed by thematic structuring.

\begin{figure*}[t]
  \centering
  \includegraphics[width=\linewidth]{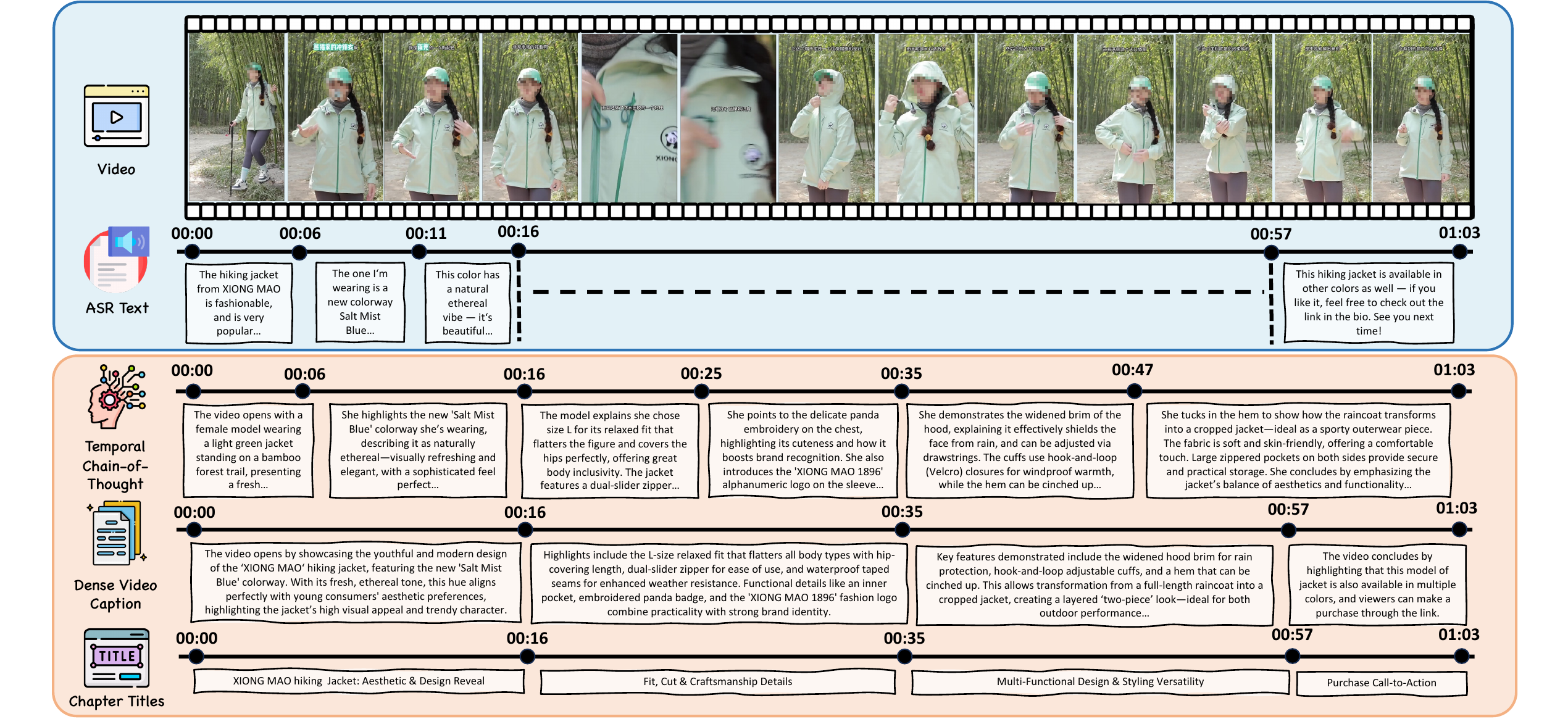}
  \caption{An annotation example from E-HVC. The model takes video frames and ASR transcripts as input, and generates hierarchical outputs: event-level Temporal Chain-of-Thought, followed by chapter-based Dense Video Captions with corresponding chapter titles.}
  \label{fig:dataset_example}
\end{figure*}

To support this task, we construct \textbf{E-HVC}, a large-scale dataset with hierarchical annotations for e-commerce videos. Each video is annotated at \textbf{two granularity levels}: (1) \textbf{Temporal Chain-of-Thought}, providing time-stamped, segment-by-segment descriptions of visual details and actions; and (2) \textbf{Chapter Summary}, organizing content into contiguous chapters with titles and concise summaries. Unlike existing datasets~\cite{li2019video,yang2024synchronized} that provide only flat narrations, E-HVC explicitly captures the reasoning process from visual grounding to high-level summarization, enabling models to learn structured narrative generation.

Building on this dataset, we develop the \textbf{HiVid-Narrator} framework with two key innovations. First, to address factual hallucinations prevalent in e-commerce videos, we integrate ASR-derived product knowledge and fine-grained visual cues into the model's reasoning process, ensuring generated narrations accurately reflect product attributes. Second, to tackle the computational inefficiency of processing information-dense frames, we propose the \textbf{S}cene-\textbf{P}rimed \textbf{A}SR-anchored Compressor (\textbf{SPA-Compressor}), a plug-and-play module that leverages ASR's natural segmentation of videos into coherent scenes. By aligning visual tokens with ASR-detected scene boundaries, SPA-Compressor compresses hundreds of redundant frame tokens into compact, scene-aware representations without sacrificing critical details.

Our main contributions are:
\begin{itemize}[leftmargin=*]
    \item We construct \textbf{E-commerce Hierarchical Video Captioning} (\textbf{E-HVC}), the first large-scale dataset with dual-granularity hierarchical annotations: event-level \textbf{Temporal Chain-of-Thought} and chapter-level \textbf{Chapter Summary}, enabling models to learn structured narration in e-commerce scenarios.
    \item We propose \textbf{SPA-Compressor}, a scene-aware video token compression module that reduces computational overhead while preserving temporal coherence, supporting flexible compression ratios up to 90\%+ token reduction, with 82.59\% achieving an optimal balance between efficiency and performance.
    
    \item Our \textbf{HiVid-Narrator} framework sets new state-of-the-art results on dense video caption, demonstrating superior factual accuracy and narrative structure compared to existing MLLMs.
\end{itemize}

%% file: sec/2_relatedwork.tex
\section{Related Work}
\begin{figure*}[t]
  \centering
  \includegraphics[width=\linewidth]{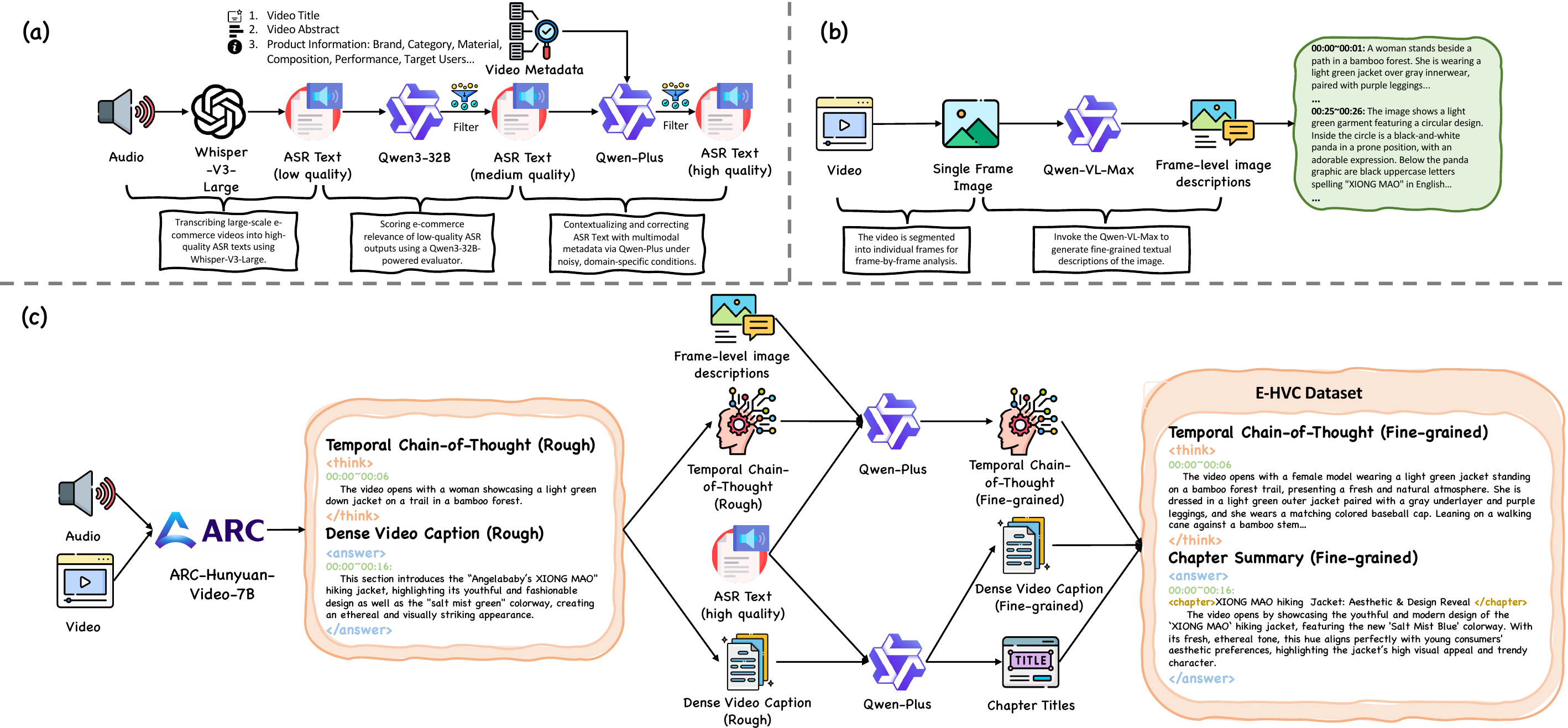}
  \caption{Multi-stage annotation pipeline for E-HVC-146K. (a) Multi-level ASR text quality enhancement; (b) Temporally aligned frame-level description generation; (c) Hierarchical reasoning from coarse to fine-grained annotations.}
  \label{fig:pipeline}
\end{figure*}
\subsection{Multi-modal Large Language Models}
Recent advances in time-sensitive multimodal large language models (MLLMs) have introduced diverse strategies to enhance temporal understanding in videos. TimeChat~\cite{ren2024timechat} adopts a dual Q-Former architecture for timestamp embedding and long-video processing via sliding windows, while VTimeLLM~\cite{huang2024vtimellm} leverages a three-stage training pipeline to progressively align visual-text semantics, detect event boundaries, and perceive timestamps. Momentor~\cite{qian2024momentor} focuses on fine-grained temporal reasoning through continuous time tokens and segment-level modeling. LITA~\cite{huang2024lita} introduces time tokens based on relative video duration and a SlowFast token architecture to enable accurate temporal localization. TRACE~\cite{guotrace} proposes a causal event modeling framework that structures video outputs into sequences of events—each composed of timestamps, salient scores, and captions—and employs task-interleaved decoding to align with the inherent structure of videos. VTG-LLM~\cite{guo2025vtg} integrates absolute-time tokens with slot-based compression to mitigate quantization errors and context constraints.

\subsection{Video Token Compression}

Video token compression methods address computational challenges through four primary approaches. \textbf{Transformation-based} methods employ pooling~\cite{xu2024pllava} or 3D convolution~\cite{cheng2024videollama} to downsample spatiotemporal features. \textbf{Similarity-based} approaches leverage temporal redundancy via clustering~\cite{jin2024chat} or optimization-based merging. \textbf{Attention-based} methods exploit visual attention sparsity to prune tokens, with FastV~\cite{chen2024image} removing 50\% tokens after shallow layers and VisionZip~\cite{yang2025visionzip} retaining high-attention tokens before clustering. \textbf{Query-based} techniques use learned queries for token distillation (e.g., BLIP-3-Video~\cite{ryoo2024xgen} abstracts frames into 16-32 tokens) or cross-modal selection~\cite{shenlongvu}.

\subsection{Video Narration Generation}
Video narration generation datasets have evolved from basic captioning to structured storytelling annotations. Early benchmarks like YouCook2~\cite{zhou2018towards}, ActivityNet Captions~\cite{krishna2017dense}, and Charades-STA~\cite{gao2017tall} provided temporal event descriptions but lacked narrative coherence. Video Storytelling dataset by Li et al.~\cite{li2019video} introduced coherent narratives across sequential clips, while E-SyncVidStory~\cite{yang2024synchronized} extended this with synchronized narrations and knowledge annotations for e-commerce videos. However, existing datasets treat narration as flat text without capturing the reasoning process from visual grounding to high-level summarization. Moreover, e-commerce video datasets remain limited, with most focusing on general domains that miss domain-specific challenges such as product detail verification and fast-paced scene transitions. Our E-HVC dataset addresses these gaps by providing hierarchical annotations with explicit reasoning traces and chapter-based summaries, enabling models to learn both fine-grained visual grounding and structured narrative generation in information-dense e-commerce contexts.

%% file: sec/3_method.tex
\section{HiVid-Narrator}
\begin{figure*}[t]
  \centering
  \includegraphics[width=\linewidth]{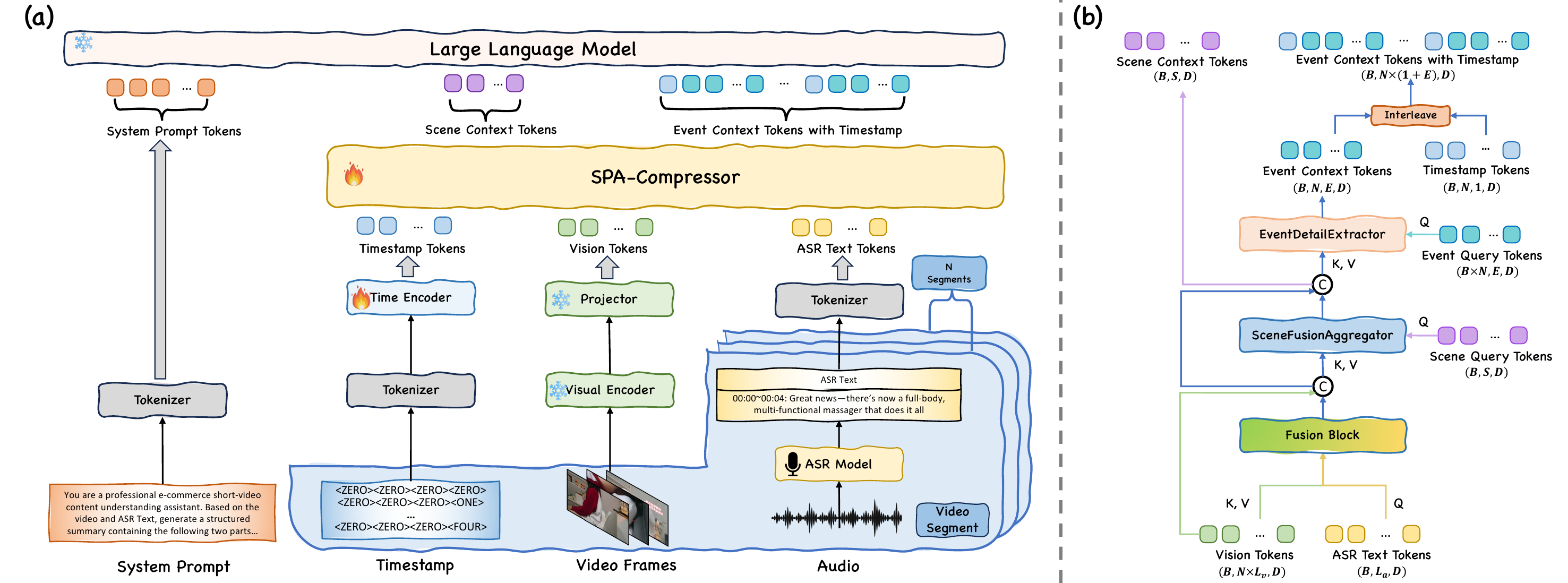}
\caption{(a) Architecture of HiVid-Narrator. The model processes four types of inputs: system prompt tokens, timestamp tokens, vision tokens, and ASR text tokens. These are compressed by the SPA-Compressor module before being fed into the LLM along with scene context tokens to generate event context tokens with timestamps. (b) Architecture of SPA-Compressor. The module consists of three stages: Fusion Block fuses vision and ASR tokens, SceneFusionAggregator extracts scene-level context via cross-attention with scene query tokens, and EventDetailExtractor generates event-level representations with timestamp tokens through transformer decoder blocks.}
  \label{fig:arch}
\end{figure*}

\subsection{Annotation Pipeline for E-HVC-146K}

To construct the E-HVC dataset, we collect over 5 million product recommendation videos from a major e-commerce platform, where content creators demonstrate and review products across diverse categories. We design a multi-stage annotation pipeline that progressively enhances data quality through hierarchical refinement while filtering low-quality content at each stage, as illustrated in Figure~\ref{fig:pipeline}. The pipeline consists of three key stages: (1) multi-level quality enhancement for ASR text construction that filters noisy segments via relevance scoring and corrects domain-specific terminology, (2) temporally aligned frame-level image description generation that captures fine-grained visual details, and (3) hierarchical reasoning-based dense video caption generation that synthesizes multimodal information into structured Temporal Chain-of-Thought and Chapter Summary through coarse-to-fine refinement. This progressive quality control yields 146K training videos and 1,852 evaluation videos (E-HVC-Bench).

\subsubsection{Multi-level Pipeline for ASR Text Construction} 
E-commerce videos contain crucial product attribute descriptions in speech, such as materials, ingredients, and performance parameters, which are often difficult to observe directly from visual frames. However, directly using general-purpose speech recognition models presents significant challenges: lack of e-commerce-specific discourse understanding, background music interference, low accuracy for brand names and model numbers, recognition degradation from regional accents and fast speech.

We address these issues through a three-stage pipeline. First, Whisper-V3-Large generates initial ASR transcriptions from raw audio. Second, a Qwen3-32B-based relevance evaluator filters text segments using few-shot prompting to identify e-commerce content (product introductions, features, pricing) and assigns quality scores from 0-5, with a threshold of 4 filtering out background noise and low-quality segments. Finally, we perform contextualized correction by encoding video metadata (titles, summaries, structured product attributes including brand, category, materials, ingredients) into JSON format and inputting them with the initial ASR text into Qwen-Plus to correct domain-specific errors.

\subsubsection{Frame-Level Image Descriptions} E-commerce videos are information-dense and fast-paced, requiring detailed capture of product attributes at fine temporal granularity. To complement video models' limited capability in perceiving static visual details, we sample videos at one frame per second with precise timestamps and invoke Qwen-VL-Max to generate detailed frame-level descriptions. We design specialized prompts guiding the model to focus on product physical attributes, spatial arrangements, textual information, and usage demonstrations. These fine-grained static observations provide essential visual grounding that video-native models often overlook, serving as crucial input for subsequent dense video captioning.

\subsubsection{Hierarchical Reasoning for Dense Video Caption Dataset Generation}

To organize fragmented multimodal information into structured video narratives, we divide dataset construction into two stages: coarse-grained initial generation and fine-grained refinement. In the coarse-grained stage, we use ARC-Hunyuan-Video-7B~\cite{ge2025arc}, which takes video audio streams as input, and outputs an initial temporal chain-of-thought and rough chapter summaries, establishing the overall temporal framework.

The fine-grained refinement stage processes the temporal chain-of-thought and chapter summaries separately. For temporal chain-of-thought generation, we fuse four types of information into Qwen-Plus: frame-level image descriptions (providing fine-grained visual details to compensate for ARC-Hunyuan-Video's limited visual understanding), high-quality ASR text (providing accurate speech information), video metadata (providing global context), and the initial chain-of-thought from ARC-Hunyuan-Video (providing overall temporal framework). The model outputs a detailed temporal chain-of-thought that chronologically records key observations for each segment and the correspondence between visual and audio information. 

For chapter summary generation, we input the refined temporal chain-of-thought, high-quality ASR text, video metadata, and initial chapter summaries from ARC-Hunyuan-Video into Qwen-Plus. The model outputs structured chapter summaries, where each chapter contains three elements: time span, chapter title, and content summary. This step-by-step generation approach ensures consistency between the temporal chain-of-thought and chapter summaries while avoiding information omission and logical gaps common in single-step generation.For the training set, we apply rule-based filtering to remove videos with excessively long or short chapter durations. For E-HVC-Bench, we conduct manual inspection to verify the reasonableness of temporal segmentation and chapter boundaries. More details are provided in Appendix.

\subsection{Overview of HiVid-Narrator}
Long video narration requires processing extended temporal sequences while maintaining both global coherence and fine-grained details. We propose HiVid-Narrator, a hierarchical compression framework that leverages speech alignment to structure multi-modal video representations. Our approach first organizes video frames and ASR transcripts into sentence-aligned segments, creating structured input sequences where each segment is anchored by timestamp markers. The compression is then performed by our SPA-Compressor, which progressively aggregates information through scene-level global summarization and event-level temporal extraction. Here, \textit{scenes} capture video segment semantics (e.g., product introduction, feature demonstration), while \textit{events} represent frame-specific details (e.g., specific gestures, visual attributes). This produces compact hierarchical representations that enable efficient processing of videos.

\subsection{Sentence-Level Interleaved Input Sequence}
We construct the input sequence by interleaving visual and textual information at the sentence granularity. Since a single semantic event (e.g., "introducing product features") typically comprises multiple ASR sentences describing different attributes, sentence-level interleaving naturally achieves fine-grained audio-visual alignment without requiring explicit event boundary detection. Each frame is represented by a timestamp token followed by its visual tokens, with ASR sentence tokens inserted immediately after their temporally corresponding frames.
\textbf{Formal Definition.} 
Given $N$ frames and $M$ ASR sentences, the input sequence is constructed as:
\begin{equation}
\mathcal{X} = [T_0, V_0, T_1, V_1, S_1,\ldots, T_{N-1}, V_{N-1}, S_M]
\end{equation}
where $T_i \in \mathbb{R}^d$ denotes the timestamp token for frame $i$, $V_i \in \mathbb{R}^{L_v \times d}$ denotes the visual tokens extracted from frame $i$, and $S_j \in \mathbb{R}^{L_s \times d}$ denotes the tokens from the $j$-th ASR sentence.

\subsection{Scene-Primed ASR-anchored Compressor}

E-commerce videos present a critical computational challenge for multimodal large language models. More visual tokens provide richer visual information and significantly improve model performance, but due to the $\mathcal{O}(n^2)$ complexity of transformer self-attention~\cite{vaswani2017attention}, a large number of visual tokens cause explosive growth in computational cost. Dense frame sampling combined with high-resolution patch encoding results in vision tokens that dominate the input sequence, seriously affecting both training efficiency and inference practicality. While ASR transcripts are relatively short, they provide complementary semantic cues about product features and demonstration intent that can guide more effective visual token compression.

We propose SPA-Compressor, a learnable module that compresses multimodal tokens into a hierarchical representation $\mathbf{H} \in \mathbb{R}^{B \times (S + N(1+E)) \times D}$, where $S$ scene tokens capture global context and $E$ event tokens per frame preserve temporal details. This design achieves drastic token reduction while maintaining task-critical information through four specialized components: Vision-ASR cross-modal fusion enriches visual understanding with spoken semantic cues, SceneFusionAggregator distills holistic video understanding into compact scene queries, EventDetailExtractor preserves frame-specific details with temporal anchoring, and hierarchical token assembly implements a global-then-local information flow.
\subsubsection{Vision-ASR Cross-Modal Fusion}

We introduce explicit Vision→ASR cross-attention to enrich spoken descriptions with concurrent visual information.

Given normalized ASR embeddings and flattened vision tokens, we compute visually-grounded ASR representations through cross-attention followed by feed-forward transformation:
\begin{equation}
\mathbf{A}' = \bar{\mathbf{A}} + \text{Cross-Attn}(\mathbf{Q}{=}\bar{\mathbf{A}}, \mathbf{K}{=}\mathbf{V}_f, \mathbf{V}{=}\mathbf{V}_f)
\end{equation}
\begin{equation}
\mathbf{A}_{\text{fused}} = \mathbf{A}' + \text{FFN}(\text{LN}(\mathbf{A}'))
\end{equation}
where $\bar{\mathbf{A}} = \text{LN}(\mathbf{A}) \in \mathbb{R}^{B \times L_a \times D}$ and $\mathbf{V}_f = \text{Flatten}(\text{LN}(\mathbf{V})) \in \mathbb{R}^{B \times N \cdot L_v \times D}$, with LN denoting layer normalization. The resulting $\mathbf{A}_{\text{fused}}$ associates spoken phrases with specific visual regions, enabling downstream modules to leverage multimodal semantics for compression.

\subsubsection{SceneFusionAggregator: Global Context Extraction}

E-commerce videos naturally contain scene-level semantics spanning the entire duration—product category, setting, and narrative arc. This holistic understanding is essential for generating coherent chapter summaries but is diluted when scattered across thousands of vision tokens. We introduce learnable scene queries as an information bottleneck to aggregate global context.

To further extract visual information, We concatenate fused ASR and vision tokens as input for scene aggregation:
\begin{equation}
\mathbf{M}_s = [\mathbf{A}_{\text{fused}}, \mathbf{V}_f] \in \mathbb{R}^{B \times (L_a + N \cdot L_v) \times D}
\end{equation}

Scene queries $\mathbf{Q}_s \in \mathbb{R}^{B \times S \times D}$ are randomly initialized learnable embeddings that attend to $\mathbf{M}_s$ through $L_s$ cross-attention layers. Each layer applies cross-attention followed by feed-forward transformation:
\begin{align}
\mathbf{H}_s^{(\ell)} &= \mathbf{H}_s^{(\ell-1)} + \text{Cross-Attn}(\text{LN}(\mathbf{H}_s^{(\ell-1)}), \mathbf{M}_s) \\
\mathbf{H}_s^{(\ell)} &= \mathbf{H}_s^{(\ell)} + \text{FFN}(\text{LN}(\mathbf{H}_s^{(\ell)})) \quad \text{for } \ell=1,\ldots,L_s
\end{align}

The final scene representation is $\mathbf{H}_{\text{scene}} = \mathbf{H}_s^{(L_s)} \in \mathbb{R}^{B \times S \times D}$, where $\mathbf{H}_s^{(0)} = \text{LN}(\mathbf{Q}_s)$. By attending to both vision and ASR simultaneously across all timestamps, scene queries distill multimodal characteristics into compact tokens that provide the LLM with global context necessary for generating chapter-based summaries.

\subsubsection{EventDetailExtractor: Temporal Detail Preservation}

While scene queries capture global understanding, dense captioning requires frame-specific details with temporal grounding. We leverage transformer decoder architecture where event queries extract fine-grained information anchored by timestamp encodings. The key design choice is to include scene tokens in the cross-attention context, implementing scene-primed reasoning—events are interpreted within global context rather than in isolation.

We construct the cross-attention context by concatenating fused ASR and scene representations:
\begin{equation}
\mathbf{M}_e = [\mathbf{A}_{\text{fused}}, \mathbf{H}_{\text{scene}}] \in \mathbb{R}^{B \times (L_a + S) \times D}
\end{equation}

Event queries $\mathbf{Q}_e \in \mathbb{R}^{(B \cdot N) \times E \times D}$ are learnable embeddings replicated for all $N$ frames and processed through $L_e$ transformer decoder layers:
\begin{align}
\mathbf{H}_e^{(\ell)} &= \mathbf{H}_e^{(\ell-1)} + \text{Self-Attn}(\text{LN}(\mathbf{H}_e^{(\ell-1)})) \\
\mathbf{H}_e^{(\ell)} &= \mathbf{H}_e^{(\ell)} + \text{Cross-Attn}(\text{LN}(\mathbf{H}_e^{(\ell)}), \mathbf{M}_e) \\
\mathbf{H}_e^{(\ell)} &= \mathbf{H}_e^{(\ell)} + \text{FFN}(\text{LN}(\mathbf{H}_e^{(\ell)})) \quad \text{for } \ell=1,\ldots,L_e
\end{align}

where $\mathbf{H}_e^{(\ell)} \in \mathbb{R}^{(B \cdot N) \times E \times D}$ and $\mathbf{H}_e^{(0)} = \text{LN}(\mathbf{Q}_e$). The self-attention enables event queries to specialize across different aspects, while cross-attention retrieves relevant information from ASR and scene tokens. The final event representation is $\mathbf{H}_{\text{event}} = \mathbf{H}_e^{(L_e)}$.

\subsubsection{Hierarchical Token Assembly}

We construct the final compressed sequence by interleaving timestamps with events, then concatenating with scene tokens. This organization implements a global-then-local information flow: the LLM first processes holistic scene context, then interprets frame-specific details within that context—mirroring the coarse-to-fine reasoning required for chapter-based summarization.

For each frame, we interleave its timestamp encoding with corresponding event tokens to create explicit temporal boundaries. This interleaving ensures that the LLM can ground each event description to its precise occurrence time, which is critical for generating temporally-accurate dense captions. The interleaved frame-level tokens are then concatenated with global scene tokens:
\begin{equation}
\mathbf{H}_{\text{frame}}^{(i)} = [\mathbf{T}_e^{(i)}, \mathbf{H}_{\text{event}}^{(i)}] \in \mathbb{R}^{B \times (1+E) \times D}
\end{equation}
\begin{equation}
\mathbf{H}_{\text{final}} = [\mathbf{H}_{\text{scene}}, \mathbf{H}_{\text{frame}}^{(1)}, \ldots, \mathbf{H}_{\text{frame}}^{(N)}]
\end{equation}

where $\mathbf{T}_e^{(i)} \in \mathbb{R}^{B \times 1 \times D}$ is the timestamp token from the time encoder for the $i$-th frame, $\mathbf{H}_{\text{event}}^{(i)} \in \mathbb{R}^{B \times E \times D}$ denotes the event tokens for the $i$-th frame, and $\mathbf{H}_{\text{final}} \in \mathbb{R}^{B \times (S + N(1+E)) \times D}$. This produces a token sequence where scene tokens precede all frame-level tokens, and within each frame, the timestamp token immediately precedes its associated event tokens, achieving drastic compression while preserving hierarchical task structure through explicit scene/event separation and timestamps.

\begin{table*}[t]
    \centering
    \small
    \setlength{\tabcolsep}{5pt}
    \begin{tabular}{lccccccc}
        \toprule
        \multirow{2}{*}{\textbf{Models}} & \multicolumn{3}{c}{\textbf{YouCook2}} & \multicolumn{3}{c}{\textbf{ActivityNet Captions}} \\
        \cmidrule(lr){2-4} \cmidrule(lr){5-7}
        & \textbf{SODA\_c} & \textbf{CIDEr} & \textbf{F1} & \textbf{SODA\_c} & \textbf{CIDEr} & \textbf{METEOR}  \\
        \midrule
        Valley~\cite{luo2023valley} (arXiv 2023) & 0.1 & 0.0 & 1.5 & 0.3 & 1.8 & 0.8 \\
        VideoChat~\cite{li2023videochat} (arXiv 2023)& 0.2 & 0.6 & 3.4 & - & - & - \\
        VTimeLLM\cite{huang2024vtimellm} (CVPR 2024) & 1.0 & 3.6 & 9.1 & 5.8 & \textbf{ 27.6} & 6.8 \\
        Monmentor~\cite{qian2024momentor} (arXiv 2024) & - & - & - & 2.3 & 14.9 & 4.7 \\
        TimeChat~\cite{ren2024timechat} (CVPR 2024) & 1.2 & 3.4 & 12.6 & 4.7 & 19.0 & 5.7 \\
        VTG-LLM~\cite{guo2025vtg} (AAAI 2025) & 1.5 & 5.0 & 17.5 & 5.1 & 20.7 & 5.9 \\
        TRACE~\cite{guotrace} (ICLR 2025) & 2.2 & 8.1 & 22.4 & 6.0 & 25.9 & 6.4 \\
        \midrule
        HiVid-Narrator w/o SPA-Compressor & 2.1 & 6.7 & 25.1 & 6.0 & 24.4 & 6.7 \\
        HiVid-Narrator w/ SPA-Compressor & \textbf{2.4} & \textbf{8.4} & \textbf{30.5} & \textbf{6.9} & 26.6 & \textbf{7.8} \\
        \bottomrule
    \end{tabular}
    \caption{Comparison on YouCook2 (zero-shot) and ActivityNet Captions datasets.}

\label{tab:main}
\vspace{-5pt}
\end{table*}
\subsection{Training Strategy}

We adopt a two-stage training strategy to progressively equip the model with both task-specific capabilities and compression efficiency while maintaining performance.

\noindent\textbf{Stage 1: Task Adaptation.}
In the first stage, we train the baseline model without the SPA-Compressor module. The objective of this stage is to transfer the pre-trained capabilities to our target task domain. The model learns to process multi-modal inputs (video, text, and timestamp) and generate accurate event localization and captioning outputs.

\noindent\textbf{Stage 2: Compression-Aware Fine-tuning.}
In the second stage, we integrate the SPA-Compressor module and the GRU-based Time Encoder into the architecture and continue training. This stage aims to enable the model to maintain its localization and captioning performance while operating with significantly compressed token sequences. The compression module learns to extract and preserve task-critical information from the original long-context inputs, while the Time Encoder learns to generate precise timestamp embeddings for temporal grounding.



%% file: sec/4_experiments.tex
\section{Experiments}

\subsection{Experiment Setup}

\subsubsection{Dataset}

We train our model on a diverse collection of dense video captioning datasets to ensure broad coverage of temporal understanding tasks. The training corpus comprises five datasets: ActivityNet Captions~\cite{krishna2017dense}, COIN~\cite{tang2019coin}, ViTT~\cite{huang2020multimodal}, VTG-IT~\cite{guo2025vtg}, and our proposed E-HVC-146K. For all datasets, we use Whisper-V3-Large to extract sentence-level ASR texts with corresponding time spans. E-HVC-146K undergoes our multi-stage annotation pipeline, while other datasets retain their original annotations. For the two-stage training, we split the mixed corpus into two equal subsets (1:1 ratio), with E-HVC-146K maintaining consistent product category distribution across both stages to prevent domain shift.

\subsubsection{Benchmark and Evaluation Metrics}

We evaluate our approach on three benchmarks with standard dense video captioning metrics. For each benchmark, we use Whisper-V3-Large to generate sentence-level ASR Texts with corresponding time spans.

\begin{figure}[h]
  \centering
  \includegraphics[width=\linewidth]{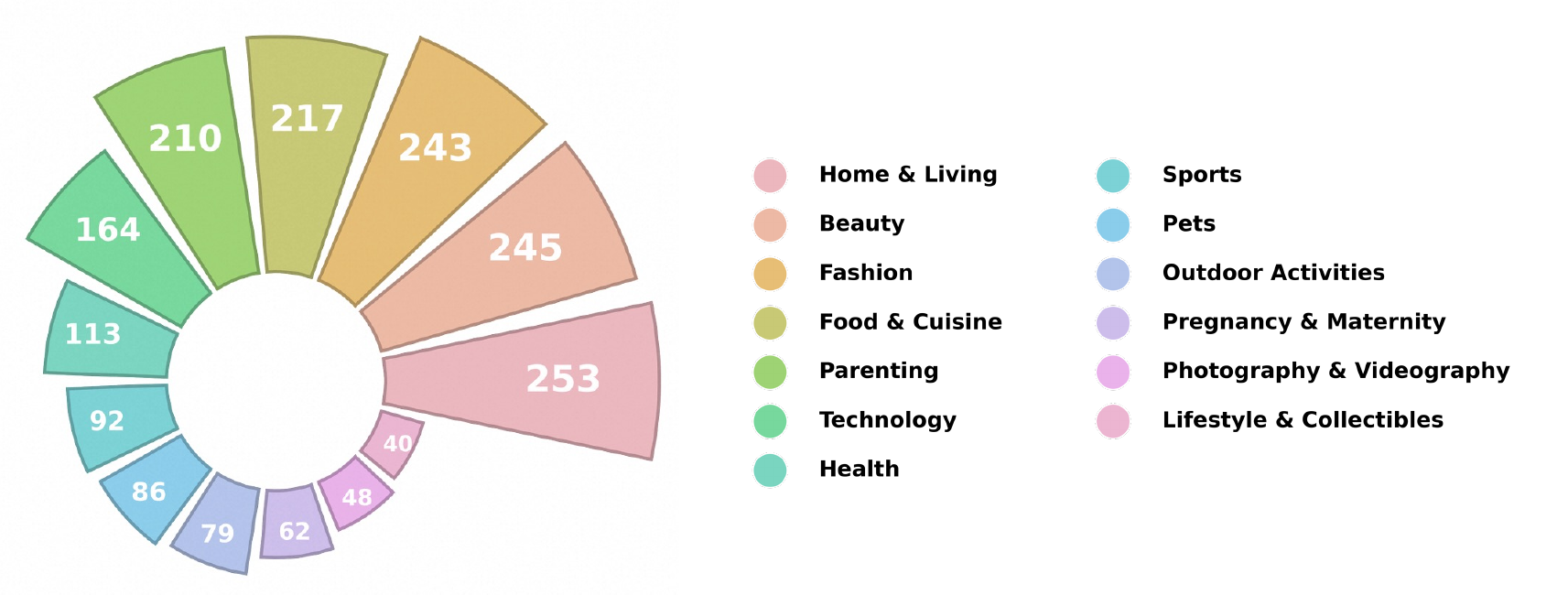}
  \caption{E-HVC-Bench: Distribution of 1,852 benchmark videos across 13 categories.}
  \label{fig:bench}
\end{figure}

\noindent\textbf{YouCook2} is evaluated in a zero-shot setting, reporting SODA\_c~\cite{fujita2020soda}, CIDEr~\cite{vedantam2015cider}, and F1 score for temporal localization accuracy.

\noindent\textbf{ActivityNet Captions} evaluation employs SODA\_c, CIDEr, and METEOR~\cite{banerjee2005meteor} to assess both content quality and temporal precision.

\noindent\textbf{E-HVC-Bench} comprises 1,852 videos spanning 13 categories (Home \& Living, Beauty, Fashion, Food \& Cuisine, Parenting, Technology, etc.), as shown in Figure~\ref{fig:bench}. We report SODA\_c, CIDEr, and METEOR for dense caption quality, with an additional BERTScore~\cite{zhang2019bertscore} specifically for chapter title generation to capture semantic similarity under paraphrasing variations common in product descriptions.

\subsubsection{Implementation Details}

We build upon Keye-VL~\cite{team2025kwai}, initializing our projector and LLM with its pretrained weights. We use SigLIP-400M-384-14~\cite{zhai2023sigmoid} as the vision encoder and Qwen3-8B~\cite{yang2025qwen3} as the LLM. Training proceeds in two stages with a total batch size of 64 distributed across 16 NVIDIA H200 GPUs. We employ the AdamW optimizer~\cite{loshchilov2017decoupled}.

In Stage 1, we train for 1 epoch with learning rate $1 \times 10^{-4}$, which undergoes full fine-tuning to adapt to dense captioning outputs while preserving pretrained vision-language representations. This stage requires 23 hours. Stage 2 trains for 1 epoch with learning rate $5 \times 10^{-5}$, freezing all other components to learn hierarchical token compression without disrupting the aligned multimodal backbone. This stage completes in 14 hours. For the SPA-Compressor architecture, we set the number of scene tokens $S{=}64$, event tokens per frame $E{=}32$, scene aggregator layers $L_s{=}2$, and event extractor layers $L_e{=}2$. Video inputs are sampled at $N{=}144$ frames.

\subsection{Main Results}
\noindent\textbf{Performance on Public Benchmarks.}
Table~\ref{tab:main} presents zero-shot evaluation on YouCook2 and supervised results on ActivityNet Captions. On YouCook2, HiVid-Narrator with SPA-Compressor achieves substantial improvements over the previous state-of-the-art TRACE, with F1 score reaching 30.5. This demonstrates superior temporal localization accuracy, validating our hierarchical architecture's ability to precisely identify event boundaries through dual-granularity scene-event modeling.

On ActivityNet Captions, our method achieves 7.8 METEOR, outperforming all baselines by significant margins, indicating that fine-grained frame descriptions combined with hierarchical aggregation generate more accurate captions aligned with ground truth annotations.

Comparing configurations with and without SPA-Compressor reveals consistent improvements across both datasets. The hierarchical compression boosts SODA\_c by 14.3\% on YouCook2 (2.1→2.4) and 15.0\% on ActivityNet Captions (6.0→6.9), demonstrating that scene-event decomposition effectively captures multi-granularity temporal structures while reducing computational overhead through token compression.

\begin{table}[h]
    \centering
    \small
    \setlength{\tabcolsep}{1.5pt}
    \begin{tabular}{lcccc}
        \toprule
        \textbf{Models} & \textbf{SODA\_c} & \textbf{CIDEr} & \textbf{METEOR} & \textbf{BERTScore} \\
        \midrule
        \multicolumn{5}{l}{\textit{Zero-shot evaluation:}} \\
        Qwen2.5-VL(7B)~\cite{bai2025qwen2} & 2.91 & 0.59 & 13.29 & 65.82 \\
        Qwen2.5-VL(32B)~\cite{bai2025qwen2} & 3.22 & 1.21 & 20.26 & 67.11 \\
        Qwen2.5-VL(72B)~\cite{bai2025qwen2} & 3.31 & 1.24 & 22.21 & 67.20 \\
        Gemini 2.5 Flash~\cite{comanici2025gemini} & 4.03 & 1.25 & 18.51 & 63.48 \\
        Gemini 2.5 Pro~\cite{comanici2025gemini} & 4.30 & 1.41 & 22.35 & 66.58 \\
        GPT-4o~\cite{hurst2024gpt} & 2.38 & 1.07 & 21.29 & 67.73 \\
        \midrule
        InternVL3(8B) & 11.98 & 1.04 & 27.38 & 67.75 \\
        Qwen2.5-VL(7B) & 12.27 & 1.12 & 27.74 & 68.83 \\
        Keye-VL$^\dagger$~\cite{team2025kwai} & 12.83 & 1.34 & 28.29 & 69.23 \\
        HiVid-Narrator$^\dagger$ & & & & \\
        \quad w/o SPA-Compressor & 13.94 & \textbf{1.46} & 31.32 & 68.34 \\
        \quad w/ SPA-Compressor & \textbf{14.48} & 1.45 & \textbf{32.01} & \textbf{74.25} \\
        \bottomrule
    \end{tabular}
    \caption{Comparison on E-HVC-Bench. $^\dagger$ indicates models fine-tuned on E-HVC-146K training set. HiVid-Narrator uses S=64 scene tokens and E=32 event tokens. Zero-shot baselines establish task difficulty while fine-tuned results demonstrate in-domain performance. More results and qualitative visualizations can be found in the Appendix.}
\label{tab:ehvc_results}
\vspace{-5pt}
\end{table}

\noindent\textbf{Performance on E-HVC-Bench.}
Table~\ref{tab:ehvc_results} evaluates methods on E-HVC-Bench, where zero-shot MLLMs achieve only 2.38-4.30 SODA\_c, demonstrating task difficulty, while fine-tuned models show substantial improvements: our HiVid-Narrator achieves 14.48 SODA\_c, outperforming Keye-VL (12.83) by 12.9\%, validating our hierarchical scene-event architecture and SPA-Compressor design. Without SPA-Compressor, our model achieves 13.94 SODA\_c, validating the effectiveness of our two-stage captioning paradigm.

\begin{table}[h]
    \centering
    \small
    \setlength{\tabcolsep}{2pt}
    \begin{tabular}{lcccc}
        \toprule
        \textbf{Training Data} & \textbf{SODA\_c} & \textbf{CIDEr} & \textbf{METEOR} & \textbf{BERTScore} \\
        \midrule
        w/o TCoT & 13.16 & 1.33 & 31.21 & 68.43 \\
        w/ Rough TCoT & 13.64 & 1.36 & 31.82 & 72.25 \\
        w/ Refined TCoT & \textbf{14.48} & \textbf{1.45} & \textbf{32.01} & \textbf{74.25} \\
        \bottomrule
    \end{tabular}
    \caption{Ablation study on training data quality with Temporal Chain-of-Thought (TCoT) for HiVid-Narrator (w/ SPA-Compressor, S=64, E=32) on E-HVC-Bench.}
\label{tab:data_ablation}
\vspace{-5pt}
\end{table}

\begin{table}[h!]
    \centering
    \setlength{\tabcolsep}{1pt}
    \small
    \begin{tabular}{lcccc}
        \toprule
        \textbf{Method} & \textbf{SODA\_c} & \textbf{CIDEr} & \textbf{METEOR} & \textbf{BERTScore} \\
        \midrule
        S=8, E=32 \space \space \textit{(93.38\%)} & 12.99 & 0.92 & 30.99 & 65.26  \\
        S=16, E=32 \textit{(89.4\%)} & 13.16 & 1.11 & 31.43 & 69.11 \\
        S=32, E=32 \textit{(87.13\%)} & 13.57 & 1.32 & 31.31 & 71.90 \\
        S=64, E=32 \textit{(82.59\%)} & 14.48 &\textbf{1.45} & 32.01 & \textbf{74.25} \\
        S=64, E=8 \space \space \textit{(88.84\%)} & 13.21 & 1.08 & 31.11 & 66.24\\
        S=64, E=16 \textit{(86.76\%)} & 13.18 & 1.26 & 31.45 & 73.21 \\
        S=64, E=64 \textit{(74.26\%)} &\textbf{14.52} & 1.43  & \textbf{32.11}& 73.29 \\
        \bottomrule
    \end{tabular}
    \caption{Ablation study on query configurations. We evaluate different compression strategies and the number of scene/event queries on E-HVC-Bench. Numbers in parentheses indicate compression ratios.}
    \label{tab:ablation_compression}
\end{table}

\subsection{Ablation Study}

\noindent\textbf{Impact of Temporal Chain-of-Thought.}
Table~\ref{tab:data_ablation} demonstrates the value of hierarchical reasoning in training data construction. Rough TCoT improves SODA\_c from 13.16 to 13.64, while refined TCoT incorporating frame-level descriptions and high-quality ASR further boosts performance to 14.48, validating the effectiveness of our data pipeline.

\noindent\textbf{Impact of Hierarchical Compression.}
Table~\ref{tab:ablation_compression} shows how scene token ($S$) and event token ($E$) configurations affect performance on E-HVC-Bench. Increasing scene tokens from $S{=}8$ to $S{=}64$ (fixing $E{=}32$) yields consistent improvements: SODA\_c rises from 12.99 to 14.48 and BERTScore from 65.26 to 74.25, demonstrating that coarse-grained scene representations require sufficient capacity for long-range temporal modeling. For event tokens (fixing $S{=}64$), performance saturates at $E{=}32$—further increasing to $E{=}64$ provides minimal gains (SODA\_c: 14.48→14.52) while sacrificing compression efficiency (82.59\%→74.26\% reduction).We select $S{=}64, E{=}32$ as our default configuration, achieving 14.48 SODA\_c with 82.59\% token reduction. Notably, even aggressive compression ($S{=}8, E{=}32$, 93.38\% reduction) achieves 12.99 SODA\_c. Detailed analysis of compression ratios and their impact on performance can be found in the Appendix. 

%% file: sec/5_conclusion.tex
\section{Conclusion}
We present HiVid-Narrator, a framework for generating structured narratives from e-commerce videos through hierarchical reasoning and token compression. Our key contributions include: (1) E-HVC, a large-scale dataset with dual-granularity annotations—Temporal Chain-of-Thought and Chapter Summary—that explicitly captures the reasoning process from visual grounding to narrative structuring; (2) SPA-Compressor, a scene-aware compression module that reduces visual tokens by 82.59\% while preserving temporal coherence through hierarchical scene-event modeling. Extensive experiments demonstrate that our approach outperforms existing multimodal LLMs on video caption benchmarks, validating the effectiveness of hierarchical token organization in SPA-Compressor and the hierarchical reasoning paradigm for video narrative generation. This work establishes a foundation for video understanding systems in e-commerce applications. 

%% file: sec/X_suppl.tex
\clearpage

\setcounter{page}{1}
\maketitlesupplementary

\section{Dataset Statistics}

\begin{table*}[!htbp]
    \centering
    \small
    \setlength{\tabcolsep}{2pt}
    \begin{tabular}{lcccccc}
        \toprule
        \textbf{Dataset} & \textbf{\# Videos} & \textbf{Chapters/Video} & \textbf{Think Length/Video} & \textbf{Think Length/s} & \textbf{Answer Length/Video} & \textbf{Answer Length/s}\\
        \midrule
        YouCook2 & 2K & 7.7 & - & - & 67.7 (en) & 0.21 \\
        ActivityNet Captions & 20K & 3.7 & - & - & 47.6 (en) & 0.40 \\
        Charades-STA & 10K & 1.8 & - & - & 11.0 (en) & 0.36 \\
        ViTT & 8K & 5.0 & - & - & 110.5 (en) & 0.44 \\
        VideoStory & 20K & 6.1 & - & - & - & - \\
        Video Storytelling & 105 & 13.5 & - & - & 162.6 (en) & 0.22 \\
        E-SyncVidStory & 6K & 6.9 & - & - & 194.1 (zh) & 5.21 \\
        \midrule
        E-HVC-146K & 146K & 4.1 & 693.0 & 17.0 & 290.0(zh) & 6.94 \\
        E-HVC-Bench & 1.8K & 4.2 & 706.8 & 17.4 & 309.1(zh) & 7.42 \\
        \bottomrule
    \end{tabular}
    \caption{Comparison of video analysis datasets. For datasets without think sections, corresponding columns are marked with `-`. Text lengths are in characters; language is indicated in parentheses where applicable. The 'Chapters/Video' column represents the average number of clips per video.}
\label{tab:ehvc_results}
\vspace{-5pt}
\end{table*}

To provide a comprehensive overview of the datasets used in our study, we present detailed statistics in Table~\ref{tab:ehvc_results}. This table compares our proposed E-HVC dataset (both the large-scale training set E-HVC-146K and the evaluation benchmark E-HVC-Bench) with several existing public video analysis datasets.

The statistics include the number of videos, the average number of temporal segments or chapters per video (referred to as 'Chapters/Video', which corresponds to the average number of clips per video as defined in the main paper), and the average text length metrics. For datasets that do not feature a dual-granularity annotation scheme like E-HVC (i.e., they lack a 'Temporal Chain-of-Thought' equivalent), the 'Think Length' columns are marked with -. The 'Answer Length' columns represent the average length of the main descriptive or summary text per video and per second. For E-HVC, these correspond to the average length of the 'Chapter Summary' component. The unit for text length is characters, and the language is indicated in parentheses where known.

As detailed in the main paper, the E-HVC dataset is unique in providing dual-granularity annotations: the Temporal Chain-of-Thought for fine-grained event-level understanding and the Chapter Summary for coherent, high-level narrative structuring. This is reflected in the statistics, where E-HVC datasets have values for both 'Think' and 'Answer' length metrics, while other datasets only have 'Answer' metrics. The statistics confirm that E-HVC-146K is a large-scale dataset, and both E-HVC-146K and E-HVC-Bench contain significantly more detailed text descriptions (both in the 'Think' and 'Answer' components) compared to the baseline public datasets, reflecting the rich, structured nature of our annotations tailored for e-commerce video narrative generation.

\begin{table}[h]
    \centering
    \small
    \setlength{\tabcolsep}{1.5pt}
    \begin{tabular}{lcccc}
        \toprule
        \textbf{Models} & \textbf{SODA\_c} & \textbf{CIDEr} & \textbf{METEOR} & \textbf{BERTScore} \\
        \midrule
        InternVL3(8B) & 11.98 & 1.04 & 27.38 & 67.75 \\
        Qwen2.5-VL(7B) & 12.27 & 1.12 & 27.74 & 68.83 \\
        Keye-VL(8B) & 12.83 & 1.34 & 28.29 & 69.23 \\
        HiVid-Narrator & \textbf{14.48} & 1.45 & \textbf{32.01} & \textbf{74.25} \\
        \bottomrule
    \end{tabular}
    \caption{Comparison on E-HVC-Bench. $^\dagger$ indicates models fine-tuned on E-HVC-146K training set. HiVid-Narrator uses S=64 scene tokens and E=32 event tokens. Zero-shot baselines establish task difficulty while fine-tuned results demonstrate in-domain performance. }
\label{tab:ehvc_bench_supp}
\vspace{-5pt}
\end{table}

\section{Additional Experimental Results on E-HVC-Bench}
\label{app:exp_ehvc_bench}

Table~\ref{tab:ehvc_bench_supp} presents a detailed comparison of various models on the E-HVC-Bench evaluation set. This table supplements the results shown in the main paper (Table~\ref{tab:ehvc_results}), focusing specifically on the in-domain performance of models fine-tuned on the E-HVC-146K training set. As noted in the main text, zero-shot multimodal large language models achieve relatively low scores on this benchmark, confirming the task's difficulty. Fine-tuned models, including InternVL3, Qwen2.5-VL, and our baseline Keye-VL, show significantly improved performance. Our proposed HiVid-Narrator consistently achieves the highest scores across multiple metrics, particularly excelling in SODA\_c (14.48) and BERTScore (74.25), demonstrating its superior ability to generate temporally coherent and semantically accurate narratives for e-commerce videos compared to strong baselines.
\section{Compression Ratio Analysis for SPA-Compressor}
\label{app:compression_analysis}

This section provides a detailed analysis of the compression ratios achieved by our proposed Scene-Primed ASR-anchored Compressor (SPA-Compressor). The core idea is to compress the high-dimensional visual tokens generated by the backbone vision encoder into a more compact, hierarchical representation consisting of scene-level ($S$) and event-level ($E$) tokens.

The original input sequence for a video often contains a large number of visual tokens, which dominate the sequence and create computational overhead due to the quadratic complexity of transformer self-attention. The SPA-Compressor addresses this by leveraging ASR transcripts to segment the video into sentence-aligned segments. For each segment, it compresses the corresponding visual tokens into $S$ scene tokens and $N \times E$ event tokens, where $N$ represents the average number of image frames associated with each ASR sentence. This results in a total of $S + N \times E$ compressed visual tokens per ASR segment, significantly reducing the input length.

The compression ratio for each ASR segment is calculated as:
\begin{equation}
    \text{Compression Ratio} = \frac{S + N \times E}{N \times D_v} = \frac{S/N + E}{D_v}
\end{equation}
where $D_v$ is the dimensionality of the original visual tokens (e.g., $D_v = 384$ for siglip-so400m-patch 14-384).

The token reduction percentage is calculated as:
\begin{equation}
    \text{Token Reduction (\%)} = (1 - \text{Compression Ratio}) \times 100
\end{equation}

Our analysis of the E-HVC dataset reveals that $N$, the average number of image frames per ASR sentence, is approximately 1.836. For the specific configuration used in our main experiments, $S=64$ and $E=32$, the compression ratio is calculated as:
\begin{align}
    \text{Compression Ratio} &= \frac{64 + 1.836 \times 32}{1.836 \times 384} \\
    &= \frac{64 + 58.752}{704.832} \\
    &= \frac{122.752}{704.832} \approx 0.1741
\end{align}
This corresponds to a token reduction of $(1 - 0.1741) \times 100 \approx 82.59\%$. This aligns closely with the 82.59\% token reduction reported in the main paper (Table~\ref{tab:ablation_compression}), confirming the high efficiency of our SPA-Compressor in achieving significant input token reduction while maintaining essential visual details through the hierarchical scene-event modeling.